\documentclass[10pt,twocolumn,letterpaper]{article}

\usepackage[pagenumbers]{cvpr} 

\usepackage{graphicx}
\usepackage{amsmath}
\usepackage{amssymb}
\usepackage{booktabs}
\usepackage{amsthm}
\usepackage{algorithm}
\usepackage{algorithmic}

\usepackage[pagebackref,breaklinks,colorlinks]{hyperref}

\usepackage[capitalize]{cleveref}
\crefname{section}{Sec.}{Secs.}
\Crefname{section}{Section}{Sections}
\Crefname{table}{Table}{Tables}
\crefname{table}{Tab.}{Tabs.}

\def\cvprPaperID{} 
\def\confName{CVPR}
\def\confYear{2022}

\newtheorem{theorem}{\bf Theorem}

\begin{document}

\title{Bi-level Doubly Variational Learning for Energy-based Latent Variable Models}

\author{
Ge Kan$^1$, Jinhu L\"{u}$^1$, Tian Wang$^1$\thanks{Corresponding Author}, Baochang Zhang$^1$, Aichun Zhu$^2$\\
Lei Huang$^1$, Guodong Guo$^3$, Hichem Snoussi$^4$\\
$^1$School of Automation Science and Electrical Engineering, Institue of Artificial Intelligence, \\
Beihang University, Beijing China\\
$^2$ School of Computer Science and Technology, Nanjing Tech University, Nanjing China \\
$^3$Institute of Deep Learning, Baidu Research, \\
National Engineering Laboratory for Deep Learning Technology and Application, Beijing China\\
$^4$University of Technology of Troyes, Troyes France\\
{\tt\small \{gekan,wangtian,bczhang\}@buaa.edu.cn,jhlu@iss.ac.cn,aichun.zhu@njtech.edu.cn}\\
{\tt\small huanglei36060520@gmail.com,guoguodong01@baidu.com,hichem.snoussi@utt.fr}
}
\maketitle

\begin{abstract}
   Energy-based latent variable models (EBLVMs) are more expressive than conventional energy-based models. However,  its potential on visual tasks are limited by its training process based on  maximum likelihood estimate that  requires sampling from two intractable distributions.  In this paper, we propose Bi-level doubly variational learning (BiDVL), which is based on a new bi-level optimization framework and two tractable variational distributions to facilitate learning EBLVMs. Particularly, we lead a decoupled EBLVM consisting of a marginal energy-based distribution and a structural posterior to handle the difficulties when learning deep EBLVMs on images.  By choosing a symmetric KL divergence in the lower level of our framework, a compact BiDVL for visual tasks can be  obtained. Our model achieves impressive image generation performance over related works. It also demonstrates the significant capacity of testing image reconstruction and out-of-distribution detection. 
\end{abstract}

\section{Introduction} 

Energy-based models (EBMs) \cite{LeCun2006ATO} are known as powerful generative models widely studied in the field of machine learning, which define a general distribution explicitly by normalizing exponential negative energy. EBMs have  also  been successfully applied to solve visual tasks, such as image synthesis \cite{Du2019ImplicitGA}, classification \cite{Grathwohl2020YourCI}, out-of-distribution (OOD) detection \cite{Liu2020EnergyOOD} or semi-supervised learning \cite{Gao2020FlowCE}. Further, latent variables are incorporated to define energy-based latent variable models (EBLVMs), which are more expressive and capable of representation learning \cite{Bao2021VariationalE}. 

However, the effectiveness of EBLVM deteriorates when applying it to solve computer vision problems, because learning by maximum likelihood estimation (MLE) usually suffers from the \textit{doubly intractable} model problem \cite{Vrtes2016LearningDI}, \ie, sampling from the posterior and the joint distribution of model is nontrivial due to the intractable integrals in normalizing denominator. Recent works \cite{Du2019ImplicitGA,Nijkamp2020LearningEM} leverage gradient-based Markov Chain Monte Carlo (MCMC), such as Langevin dynamics, to approximately sample from EBMs but require lots of steps, since fewer steps may lead to arbitrarily far sampling distribution from the target. Moreover, to handle the \textit{doubly intractable} problem of EBLVMs, double MCMC sampling are needed and as a result making it infeasible on high-dimensional images. 

\begin{figure}[t]
  \centering
  \includegraphics[width=0.94\linewidth]{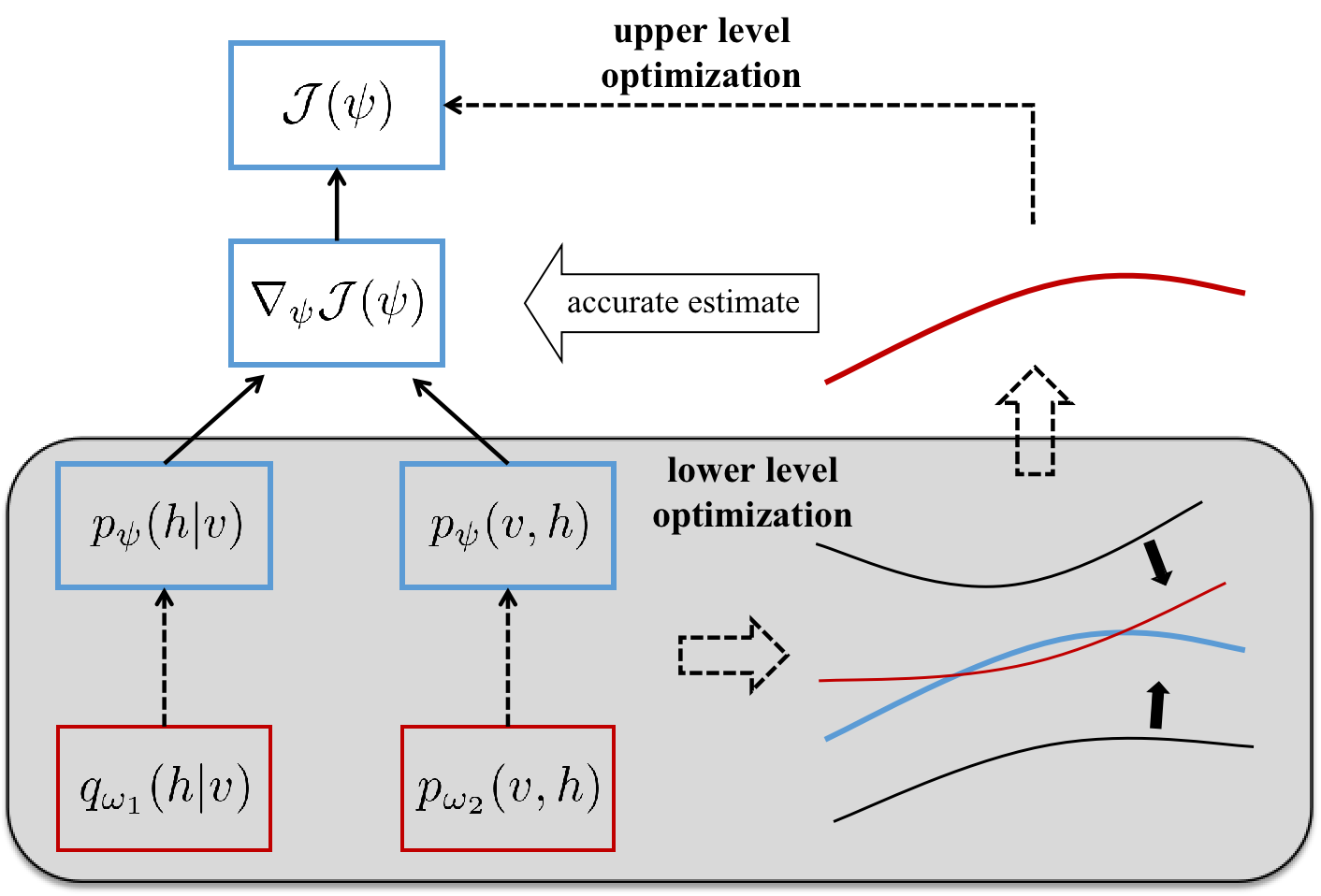}
   \caption{The motivation for BiDVL. Blue items represent the real objective optimization procedure. Red items indicate the approximate one. Real gradient, gradient estimate, and bounds are represented by blue, red and black curves, respectively. The bold red curve is the gradient estimate for upper level optimization. See \cref{sec:frame} for detailed explanation. }
   \label{fig:bidvl}
\end{figure}

To accomplish EBLVMs efficiently, this paper introduces a \textbf{Bi}-level \textbf{D}oubly \textbf{V}ariational \textbf{L}earning (\textbf{BiDVL}) framework, which is based on tractable variational posterior and variational joint distribution to estimate the gradients in MLE. We formulate BiDVL as a bi-level optimization (BLO) problem as illustrated in \cref{fig:bidvl}. In specific, the gradient estimate is compelled to fit the real one by exploring variational distributions to achieve the model distributions in the lower level, and the resulting gradient estimate is then used for optimizing the objective in the upper level. Theoretically, BiDVL is equivalent to the original MLE objective under the \textit{nonparametric assumption} \cite{Goodfellow2014GenerativeAN}, while to optimize it practically, we propose an efficient alternative optimization scheme. 

Moreover, the drawbacks of doubly variational learning come from its unstableness when learning a deep EBLVM on images. For example, AdVIL \cite{Li2020ToRY} that can be approximately degenerated from BiDVL fails to achieve a good performance. Reasons for the failure lie in: 1) A general deep EBLVM has no structural assumption about its posterior and the unconstrained posterior may transfer among diverse mode structures during training, bring about dynamic target for variational learning. 2) Direct interaction between two variational distributions is overlooked, though they all approximate the EBLVM. 3) Both the energy-based posterior and its variational approximation have no effective way of modelling the real one. Such problems may effect other methods \cite{Han2020JointTO,Bao2020BilevelSM,Bao2021VariationalE} for learning deep EBLVMs. 

To tackle these issues, we decompose the EBLVM into an energy-based marginal distribution and a structural posterior, where the latter is represented by variational posterior. Then a symmetric KL divergence is chosen in the lower level, doubly linking the variational distributions by sharing posterior in both levels, a compact BiDVL for image tasks is thus obtained. At last, the decoupled EBLVM helps to derive an importance weighted KL divergence, in which samples for Monte Carlo estimator come from dataset. By learning from data through the doubly link, the posterior models the latent space more effectively. 

Specifically, we parametrize the variational distributions with an inference model and an implicit generative model (IGM). Recently, \cite{Han2019DivergenceTF,Han2020JointTO,Xie2021LearningEM} propose to jointly train EBM, inference model and IGM. However, their formulations are derived from intuitive combination to some extent, leading to overlap and conflict among models. BiDVL, entirely for learning EBLVMs, interprets the consistency by great improvement. In summary, we make following contributions: 

1) We introduce a MCMC-free BiDVL framework for learning EBLVMs efficiently. A corresponding alternative optimization scheme is adopted to solve it. 

2) We notice several problems in learning deep EBLVMs on images. To overcome these, we define a decoupled EBLVM and choose a symmetric KL divergence in the lower level. The resulting compact objective implies the consistency for learning EBLVMs. 

3) We demonstrate the impressive image generation quality among baseline models. Besides, BiDVL also shows the remarkable performance of testing image reconstruction and OOD detection. 

\vspace{-0.1cm} 
\section{Related work}
\label{sec:related}

\noindent \textbf{Learning EBMs and EBLVMs}. As unnormalized probability model, EBMs and EBLVMs are able to fit data distribution more efficiently than other generative models. Nonetheless, they require nontrivial sampling from the model when learned by MLE. Many attempts \cite{Du2019ImplicitGA,Arbel2021GeneralizedEB} leverage MCMC to approximate the procedure but suffers from quite poor efficiency. \cite{Vrtes2016LearningDI,Bao2020BilevelSM,Bao2021VariationalE,Li2019LearningEM} avoid the nontrivial sampling based on Score Matching \cite{Hyvrinen2005EstimationON}, however need unstable and inefficient computing of score functions. Another approach is Noise Contrastive Estimation \cite{Gutmann2012NoiseContrastiveEO}, which performs worse on high-dimensional images \cite{Grathwohl2021NoMF}. Furthermore, \cite{Grathwohl2020LearningTS} proposes to learn Stein Discrepancy without sampling, \cite{Yu2020TrainingDE} generalizes MLE to a $f$-divergence variational framework, \cite{Grathwohl2021NoMF,Geng2021BoundsAA} propose MCMC-free variational approaches, while all of them aim at learning EBMs, our BiDVL accomplishes EBLVMs in a different way. 

Recent attempt \cite{Li2020ToRY} introduces two variational models to form a min-min-max nested objective. Their method is evaluated on RBM \cite{Ackley1985ALA} and DBM \cite{Salakhutdinov2009DeepBM}, but is unstable for learning deep EBLVMs on images. We however propose a compact model for stability, see \cref{sec:cobidvl} for details. 

\noindent \textbf{Joint training with other models}. Recent works also notice the compatibility between EBM and other models. \cite{Nijkamp2020LearningEM,Xiao2020ExponentialTO,Xiao2021VAEBMAS,Arbel2021GeneralizedEB,Pang2020LearningLS} utilize an EBM to exponentially tilt a generative distribution, most of them have a two-stage refinement procedure with inefficient MCMC sampling. \cite{Song2020DiscriminatorCD,Che2020YourGI} regard EBM as a discriminator of GAN, but still use MCMC to modify the generative distribution. 


\cite{Han2019DivergenceTF} proposes a \textit{divergence triangle} loss for joint training EBM, inference model and IGM, \cite{Han2020JointTO} extends it to EBLVM and \cite{Xie2021LearningEM} extends it by variational MCMC teaching \cite{Xie2018CooperativeLO}. These studies integrate KL losses intuitively, by actually assuming the EBM always fits data perfectly, which induces harmful conflict among the models involved. While our compact objective derives an importance weighted KL loss and the variational models in BiDVL are optimized simultaneously rather than their sleep-awake scheme. All the differences originate from our unified framework, showing the consistency for learning EBLVMs. 


\section{Preliminaries}
\label{sec:preli}

EBLVM defines a joint scalar energy function over visible variable $v$ and latent variable $h$. The corresponding joint distribution is
\begin{equation}
\setlength{\abovedisplayskip}{5pt}
\setlength{\belowdisplayskip}{5pt}
    p_{\psi}(v,h) = \frac{\exp{(-\mathcal{E}_{\psi}(v,h))}}{\int\exp{(-\mathcal{E}_{\psi}(v,h))}\mathrm{d}h\mathrm{d}v}, 
    \label{eq:eblvm}
\end{equation}
where $\mathcal{E}_{\psi}(v,h)$ denotes the energy function over joint space parameterized by $\psi$, $\mathcal{Z}(\psi)=\int\exp{(-\mathcal{E}_{\psi}(v,h))}\mathrm{d}h\mathrm{d}v$ is known as the \textit{partition function}. The marginal distribution is given by $p_{\psi}(v)=\int p_{\psi}(v,h)\mathrm{d}h$, while the marginal energy is $\mathcal{E}_{\psi}(v)=-\log\int\exp{(-\mathcal{E}_{\psi}(v,h))}\mathrm{d}h$. 

Training EBLVM is typically based on MLE, or equivalently minimizing the KL divergence $\mathcal{J}(\psi)=D_{\rm KL}(q(v)||p_{\psi}(v))$, where $q(v)$ denotes the empirical data distribution. As \cite{LeCun2006ATO} notes, the gradient of the negative log-likelihood objective \wrt $\psi$ can be computed by
\begin{equation}
\setlength{\abovedisplayskip}{5pt}
\setlength{\belowdisplayskip}{5pt}
\begin{gathered}
    \nabla_{\psi}\mathbb{E}_{q(v)}[-\log p_{\psi}(v)] = \nabla_{\psi}D_{\rm KL}(q(v)\Vert p_{\psi}(v)) = \\
    \mathbb{E}_{q(v)p_{\psi}(h|v)} \left[\nabla_{\psi}\mathcal{E}_{\psi}(v,h) \right]-\mathbb{E}_{p_{\psi}(v,h)} \left[\nabla_{\psi}\mathcal{E}_{\psi}(v,h) \right], 
\end{gathered} 
\label{eq:cd}
\end{equation}
where $p_{\psi}(h|v)=\frac{\exp{(-\mathcal{E}_{\psi}(v,h))}}{\int\exp{(-\mathcal{E}_{\psi}(v,h))}\mathrm{d}h}$ is the posterior distribution over latent variables. In \cref{eq:cd}, samples from $p_{\psi}(h|v)$ and $p_{\psi}(v,h)$ are required for Monte Carlo gradient estimator, which is \textit{doubly intractable} due to the integrals over the high-dimensional space. Commonly adopted MCMC however suffers from high consumption and is too bulky to be applied to both distributions simultaneously. 

\section{Learning method}
\subsection{Proposed Bi-level doubly variational learning}

To handle the \textit{doubly intractable} problem efficiently, we propose doubly variational learning, \ie, introduce two variational models to represent the variational posterior $q_{\omega_1}(h|v)$ and the variational joint distribution $p_{\omega_2}(v,h)$, where $\omega\doteq(\omega_1,\omega_2)$ is the trainable parameters. Sampling from variational distribution requires only one forward step, largely improving efficiency. 

\vspace{-0.7em}
\subsubsection{Framework}
\label{sec:frame}

In this part, we introduce the doubly variational learning under the bi-level optimization (BLO) framework. Let $\mathcal{E}_{\psi}\doteq\mathcal{E}_{\psi}(v,h)$ for clarity. We first rewrite \cref{eq:cd} as follow to show our motivation: 
\begin{gather}
\setlength{\abovedisplayskip}{5pt}
\setlength{\belowdisplayskip}{5pt}
    \nabla_{\psi}\mathcal{J}(\psi) \label{eq:split} = \mathbb{E}_{q(v)p_{\psi}(h|v)} \left[\nabla_{\psi}\mathcal{E}_{\psi} \right] - \mathbb{E}_{p_{\psi}(v,h)} \left[\nabla_{\psi}\mathcal{E}_{\psi} \right] \\
    \equiv \mathbb{E}_{q(v)q_{\omega_1}(h|v)} \left[\nabla_{\psi}\mathcal{E}_{\psi} \right] - \mathbb{E}_{p_{\omega_2}(v,h)} \left[\nabla_{\psi}\mathcal{E}_{\psi} \right] \tag{\ref{eq:split}{a}}\label{eq:splita} \\
    + \mathbb{E}_{q(v)p_{\psi}(h|v)} \left[\nabla_{\psi}\mathcal{E}_{\psi} \right] - \mathbb{E}_{q(v)q_{\omega_1}(h|v)} \left[\nabla_{\psi}\mathcal{E}_{\psi} \right] \tag{\ref{eq:split}{b}}\label{eq:splitb} \\
    + \mathbb{E}_{p_{\omega_2}(v,h)} \left[\nabla_{\psi}\mathcal{E}_{\psi} \right] - \mathbb{E}_{p_{\psi}(v,h)} \left[\nabla_{\psi}\mathcal{E}_{\psi} \right], \tag{\ref{eq:split}{c}}\label{eq:splitc}
\end{gather}
where $\equiv$ denotes the identity \wrt $\omega$, in other word, it holds whatever $\omega_1,\omega_2$ are. \Cref{eq:split} implies that, for an appropriate $\omega$ such that both terms (\ref{eq:splitb}) and (\ref{eq:splitc}) equal zero, then term (\ref{eq:splita}) is an accurate gradient estimation. 

The motivation to formulate a bi-level optimization problem is illustrated in \cref{fig:bidvl}. Variational distributions $q_{\omega_1}(h|v)$ and $p_{\omega_2}(v,h)$ (red marked items) aim to chase the model distributions $p_{\psi}(h|v)$ and $p_{\psi}(v,h)$ (blue marked items) in lower-level optimization, reducing the gap between the upper and lower bound of gradient estimate term (\ref{eq:splita}) (red curve). The gradient estimate is thus compelled to fit the real one, \ie Eq.(\ref{eq:split}) (bold blue curve). Then the estimate finally obtained (bold red curve) is used for optimizing the real objective in upper level. 

Following the formulas described in \cite{Liu2021InvestigatingBO}, we assume the solution of lower-level (LL) subproblem \wrt the upper-level (UL) variables is of a singleton, which is known as the \textit{lower-level singleton} assumption \cite{Liu2020AGF}. Therefore we define the LL subproblem as follow: 
\begin{equation}
\setlength{\abovedisplayskip}{5pt}
\setlength{\belowdisplayskip}{5pt}
    \begin{gathered}
       \omega^*(\psi) = \arg\min_{\omega\in\Omega} \mathcal{J}_{\rm LL}(\psi,\omega), \\
       \mathcal{J}_{\rm LL}(\psi,\omega) = D(q(v)q_{\omega_1}(h|v),q(v)p_{\psi}(h|v)) \\
        + D(p_{\omega_2}(v,h),p_{\psi}(v,h)),
    \end{gathered}
    \label{eq:ll}
\end{equation}
where $\Omega$ is the parameter space of variational distributions. $D$ denotes certain proper metric corresponding to the assumption on $\nabla_{\psi}\mathcal{E}_{\psi}$, since the form of terms (\ref{eq:splitb}),(\ref{eq:splitc}) reminds of the general integral probability metrics. See \cref{sec:diver} for related discussion. We should again emphasize that the solution of LL subproblem is a function \wrt the UL variables, and as a result,  $\nabla_{\psi}\mathcal{J}(\psi)=\mathbb{E}_{q(v)q_{\omega_1^*(\psi)}(h|v)} \left[\nabla_{\psi}\mathcal{E}_{\psi} \right] - \mathbb{E}_{p_{\omega_2^*(\psi)}(v,h)} \left[\nabla_{\psi}\mathcal{E}_{\psi} \right]$ holds only if $\mathcal{J}_{\rm LL}(\psi,\omega^*(\psi))=0$. Then an equivalent of term (\ref{eq:splita}) taking $\omega=\omega(\psi)$ into account is given by: 
\begin{equation}
\setlength{\abovedisplayskip}{5pt}
\setlength{\belowdisplayskip}{5pt}
    \begin{gathered}
        \mathbb{E}_{q(v)q_{\omega_1(\psi)}(h|v)} \left[\nabla_{\psi}\mathcal{E}_{\psi} \right] - \mathbb{E}_{p_{\omega_2(\psi)}(v,h)} \left[\nabla_{\psi}\mathcal{E}_{\psi} \right] \\
        = \frac{\partial D_{\rm KL}(q(v)q_{\omega_1}(h|v)\Vert p_{\psi}(v,h))}{\partial \psi}|_{\omega_1=\omega_1(\psi)} \\
        - \frac{\partial D_{\rm KL}(p_{\omega_2}(v,h)\Vert p_{\psi}(v,h))}{\partial \psi}|_{\omega_2=\omega_2(\psi)}.
    \end{gathered} \label{eq:part}
\end{equation}
Inspired by \cref{eq:part}, we define the UL subproblem as follow: 
\begin{equation}
\setlength{\abovedisplayskip}{5pt}
\setlength{\belowdisplayskip}{5pt}
    \begin{gathered}
        \psi^* = \arg\min_{\psi\in\Psi}\mathcal{J}_{\rm UL}(\psi,\omega^*(\psi)), \\
        \mathcal{J}_{\rm UL}(\psi,\omega) = D_{\rm KL}(q(v)q_{\omega_1}(h|v)\Vert p_{\psi}(v,h)) \\
        - D_{\rm KL}(p_{\omega_2}(v,h)\Vert p_{\psi}(v,h)),
    \end{gathered} \label{eq:ul}
\end{equation}
where $\Psi$ is the parameter space of energy function. Further, the gradient of UL objective \wrt $\psi$ is denoted by 
\begin{equation}
\setlength{\abovedisplayskip}{5pt}
\setlength{\belowdisplayskip}{5pt}
\begin{split}
        &\ \frac{\partial\mathcal{J}_{\rm UL}(\psi,\omega^*(\psi))}{\partial\psi}  = \frac{\partial\mathcal{J}_{\rm UL}(\psi,\omega)}{\partial\psi}|_{\omega=\omega^*(\psi)} \\
        +&\ \left( \frac{\partial\omega^*(\psi)}{\partial\psi^{\top}} \right)^{\top}\frac{\partial\mathcal{J}_{\rm UL}(\psi,\omega)}{\partial\omega}|_{\omega=\omega^*(\psi)},
\end{split}
\label{eq:ulgra}
\end{equation}
where $\frac{\partial\omega^*(\psi)}{\partial\psi}$ is known as the Best-Response (BR) Jacobian \cite{Liu2021InvestigatingBO} which captures the response of LL solution \wrt the UL variables. 

We argue that, under the \textit{nonparametric assumption} \cite{Goodfellow2014GenerativeAN}, the BLO problem (\ref{eq:ll},\ref{eq:ul}) in BiDVL is equivalent to optimize the original objective $\mathcal{J}(\psi)$. The formal theorem is presented as follow: 
\begin{theorem} \label{thm1}
Assuming that $\forall\psi\in\Psi,\exists\omega\in\Omega$ such that \\ $D(q(v)q_{\omega_1}(h|v),q(v)p_{\psi}(h|v))=0$, and \\
$D(p_{\omega_2}(v,h),p_{\psi}(v,h))=0$, then we have
\begin{equation*}
    \mathcal{J}(\psi) = \mathcal{J}_{\rm UL}(\psi,\omega^*(\psi)), \nabla_{\psi}\mathcal{J}(\psi) = \nabla_{\psi}\mathcal{J}_{\rm UL}(\psi,\omega^*(\psi)).
\end{equation*}
\end{theorem}
The \textit{nonparametric assumption} does not always hold in practice, since the variational models have limited capacity. But we can still bound the original objective as $|\mathcal{J}_{\rm UL}(\psi,\omega^*(\psi))- \mathcal{J}(\psi)|\leq C^{\prime}\cdot\mathcal{J}_{\rm LL}(\psi,\omega)$. See \cref{sec:derive,sec:proof} for detailed derivations and proof. 

\vspace{-0.7em}
\subsubsection{Alternative optimization}
\label{sec:alter}

Solving the BLO problem requires to compute the gradient of UL objective (\ref{eq:ulgra}). As \cite{Liu2021InvestigatingBO} notes, obtaining the gradient suffers from handling the BR Jacobian with a recursive derivation process. Recent attempts \cite{Bao2020BilevelSM,Metz2017UnrolledGA} leverage \textit{gradient unrolling} technique to estimate the gradient in bi-level optimization problems. Nevertheless, \textit{gradient unrolling} requires inner loop to form a recursive computation graph and more resources are allocated for backtracking. It also incurs inferior performance in our case. For efficiency, we ignore the BR Jacobian term in \cref{eq:ulgra}. Then BiDVL is optimized alternatively, \ie, when updating the parameters involved in current level, we consider the rest as fixed. Notice the missing-term alternative optimization has been widely adopted in reinforcement learning. 

\subsection{BiDVL for Image Tasks}
\label{sec:cobidvl}

BiDVL is a general framework for EBLVMs, however, doubly variational learning is usually unstable to scale to learning deep EBLVMs on image datasets. AdVIL \cite{Li2020ToRY}, which can be approximately derived from BiDVL via choosing KL divergence in the lower level, encounters the same difficulty that unstable training and inferior performance to deal with images. We attribute the failure to three key learning problems which exacerbate the instability on high-dimensional and highly multimodal image datasets:

1) The generally defined undirected EBLVM (\ref{eq:eblvm}) has no explicit structural assumption about its posterior $p_{\psi}(h|v)$ (due to the coupled joint energy represented by network), and resulting in a dynamic target, whose mode structure transfers rapidly, for $q_{\omega_1}(h|v)$ and $p_{\omega_2}(v,h)$ to chase. 

2) Though both variational distributions are to approximate $p_{\psi}(v,h)$, there is no direct interaction between them, thus misalignment may be induced in variational learning. 

3) Both $p_{\psi}(h|v)$ and $q_{\omega_1}(h|v)$ cannot model the real posterior effectively, since they learn from each other (\ref{eq:ll}) rather than from data directly. 

Next, we introduce our solutions to the problems above with BiDVL as a necessary framework, and then, a practical compact BiDVL for image tasks is consequently conducted. 

\vspace{-0.7em}
\subsubsection{Decoupled EBLVM}
\label{sec:decoup}

The problems above imply that an unconstrained posterior is confused and superfluous for modeling the latent space, which motivates us to \textit{redefined} $p_{\psi}(v)$ and $p_{\psi}(h|v)$ as an energy-based marginal distribution and a structural posterior. Considering the same parameterization of $q_{\omega_1}(h|v)$ and $p_{\psi}(h|v)$, and by $\min D(q(v)q_{\omega_1}(h|v)||q(v)p_{\psi}(h|v))$ in \cref{eq:ll}, we can directly make the energy-based posterior equal to the variational one by sharing parameters $\omega_1$, and as a consequence, we have $D(q(v)q_{\omega_1}(h|v),q(v)p_{\psi}(h|v))=0$, eliminating the confusing chase between posteriors in BiDVL as well as reducing the LL objective. We finally formulate a decoupled version of EBLVM: 
\begin{equation}
\setlength{\abovedisplayskip}{5pt}
\setlength{\belowdisplayskip}{5pt}
    \begin{gathered}
        p_{\psi^{\prime},\omega_1}(v,h)=p_{\psi^{\prime}}(v)q_{\omega_1}(h|v), \\
        p_{\psi^{\prime}}(v)\propto\exp{(-\mathcal{E}_{\psi^{\prime}}(v))}, 
    \end{gathered}
    \label{eq:jomod}
\end{equation}
where $\mathcal{E}_{\psi^{\prime}}(v)$ is the new marginal energy parametrized with $\psi^{\prime}$ and then a decoupled joint energy is \textit{redefined} by $\mathcal{E}_{\psi^{\prime},\omega_1}(v,h) = \mathcal{E}_{\psi^{\prime}}(v) - \log q_{\omega_1}(h|v)$. Note we are actually optimizing the upper bound of $\mathcal{J}(\psi)$, in other words, if the bi-level objectives are reduced, the original objective is reduced in turn, therefore we can optimize $\omega_1$ in both levels without changing the optimal solution of BiDVL. 

\vspace{-0.7em}
\subsubsection{Symmetric KL divergence}
\label{sec:symkl}

Since $\omega_1$ is optimized in both levels, a proper metric in the lower level can link variational distributions, where KL divergence is a feasible choice. However as \cite{Goodfellow2016Deep,Han2020JointTO} notice, minimizing $D_{\rm KL}(P\Vert Q)$ \wrt $Q$ compels $Q$ to cover the major support of $P$, while minimizing a reverse version $D_{\rm KL}(Q\Vert P)$, $Q$ is forced to chase the major modes of $P$. So for integrating both behavior, we choose a symmetric KL divergence, $S(P\Vert Q)=D_{\rm KL}(P\Vert Q)+D_{\rm KL}(Q\Vert P)$, in the lower level. Meanwhile, by taking the alternative optimization from \cref{sec:alter} into account, a compact BiDVL is derived from \cref{eq:ll,eq:ul}: 
\begin{equation}
\setlength{\abovedisplayskip}{5pt}
\setlength{\belowdisplayskip}{5pt}
    \begin{gathered}
        \min_{\omega} \mathcal{J}_{\rm LL}(\omega), \\
        \mathcal{J}_{\rm LL}(\omega) = D_{\rm KL}(p_{\psi^{\prime},\omega_1}(v,h)\Vert p_{\omega_2}(v,h)) \\
        +\ D_{\rm KL}(p_{\omega_2}(v,h)\Vert p_{\psi^{\prime},\omega_1}(v,h)),
    \end{gathered}
    \label{eq:newll}
\end{equation}
\begin{equation}
\setlength{\abovedisplayskip}{5pt}
\setlength{\belowdisplayskip}{5pt}
    \begin{gathered}
        \min_{\psi^{\prime},\omega_1}\mathcal{J}_{\rm UL}(\psi^{\prime},\omega_1) \\
        \mathcal{J}_{\rm UL}(\psi^{\prime},\omega_1) = D_{\rm KL}(q(v)q_{\omega_1}(h|v)\Vert p_{\psi^{\prime},\omega_1}(v,h)) \\
        -\ D_{\rm KL}(p_{\omega_2}(v,h)\Vert p_{\psi^{\prime},\omega_1}(v,h)), 
    \end{gathered} 
    \label{eq:newul}
\end{equation}
where the symmetric KL divergence in the lower level sheds light on the connection between variational distributions in two directions. Based on above solutions, we derive a weighted KL loss shown in next section, which is proven to help modelling the data latent space effectively. 

\vspace{-0.7em}
\subsubsection{Optimizing compact BiDVL} 
\label{sec:ocblo}

In this part, we present the gradients for optimizing compact BiDVL (\ref{eq:newll},\ref{eq:newul}) followed by a simplified version. Gradients are computed by: 
\begin{equation}
\setlength{\abovedisplayskip}{5pt}
\setlength{\belowdisplayskip}{5pt}
    \begin{split}
        &\ \nabla_{\omega} \mathcal{J}_{\rm LL}(\omega) \\
        =&\ \nabla_{\omega} D_{\rm KL}(p_{\psi^{\prime}}(v)q_{\omega_1}(h|v)\Vert p_{\omega_2}(v,h)) \\
        +&\ \nabla_{\omega_2} \mathbb{E}_{p_{\omega_2}(v,h)}[\mathcal{E}_{\psi^{\prime}}(v)] + \nabla_{\omega} D_{\rm KL}(p_{\omega_2}(v,h)\Vert q_{\omega_1}(h|v)) \\
        =&\ \nabla_{\omega}  \int \textcolor{red}{\frac{p_{\psi^{\prime}}(v)}{q(v)}} q(v) q_{\omega_1}(h|v)\log\frac{q(v)q_{\omega_1}(h|v)}{p_{\omega_2}(v,h)}\mathrm{d}h\mathrm{d}v \\
        +&\ \nabla_{\omega_2} \mathbb{E}_{p_{\omega_2}(v,h)}[\mathcal{E}_{\psi^{\prime}}(v)] + \nabla_{\omega} D_{\rm KL}(p_{\omega_2}(v,h)\Vert q_{\omega_1}(h|v)) \\
        =&\ \nabla_{\omega} D_{\textcolor{red}{r(v)}}(q(v)q_{\omega_1}(h|v)\Vert p_{\omega_2}(v,h)) \\
        +&\ \nabla_{\omega_2} \mathbb{E}_{p_{\omega_2}(v,h)}[\mathcal{E}_{\psi^{\prime}}(v)] + \textcolor{blue}{\nabla_{\omega} D_{\rm KL}(p_{\omega_2}(v,h)\Vert q_{\omega_1}(h|v))}, \\
    \end{split}
    \label{eq:nllgr}
\end{equation}
\begin{equation}
\setlength{\abovedisplayskip}{5pt}
\setlength{\belowdisplayskip}{5pt}
    \begin{split}
        &\ \nabla_{\psi^{\prime},\omega_1} \mathcal{J}_{\rm UL}(\psi^{\prime},\omega_1) \\
        =&\ \mathbb{E}_{q(v)}[\nabla_{\psi^{\prime}}\mathcal{E}_{\psi^{\prime}}(v)] - \mathbb{E}_{p_{\omega_2}(v,h)}[\nabla_{\psi^{\prime}}\mathcal{E}_{\psi^{\prime}}(v)] \\
        -&\ \textcolor{blue}{\nabla_{\omega_1} D_{\rm KL}(p_{\omega_2}(v,h)\Vert q_{\omega_1}(h|v))}, 
    \end{split}
    \label{eq:nulgr}
\end{equation}
where $D_{r(v)}$ denotes the importance weighted KL divergence with $r(v)=\frac{p_{\psi^{\prime}}(v)}{q(v)}$ as the importance ratio. The principal reason for utilizing importance ratio is the nontrivial sampling from $p_{\psi^{\prime}}(v)$. Notice computing the ratio is still nontrivial, we present a detailed analysis in \cref{sec:analy}. 

Importance weighted term offers additional benefits. Since $D_{\rm KL}(p_{\psi^{\prime}}(v)q_{\omega_1}(h|v)\Vert p_{\omega_2}(v,h))$ has a mode-cover behavior, $p_{\psi^{\prime}}(v)$ defined on the whole space may provide fake mode-cover information for $p_{\omega_2}(v,h)$. However, samples for estimating the weighted term come from dataset, brings $p_{\omega_2}(v,h)$ faithful mode-cover information and further helps the EBLVM faster convergence and stabler training. Furthermore, minimizing $D_{\rm KL}(q(v)q_{\omega_1}(h|v) \Vert p_{\omega_2}(v,h))$ is equivalent to optimize the \textit{evidence lower bound} of VAE, whose reconstructive behavior is proven to model the data latent space effectively. 

Minimizing $D_{\rm KL}(p_{\omega_2}(v,h)\Vert q_{\omega_1}(h|v))$ tends to increase the likelihood of generated samples under the posterior, which further enhance the alignment between visible space and latent space. By sharing the variational posterior $q_{\omega_1}(h|v)$, the decoupled EBLVM thus learns a simpler structural posterior easy to model data latent space. 

Notice the opposite terms \wrt $\omega_1$ contained in \cref{eq:nllgr,eq:nulgr} can be offset to derive simplified gradients: 
\begin{equation}
\setlength{\abovedisplayskip}{5pt}
\setlength{\belowdisplayskip}{5pt}
    \begin{gathered}
        \nabla_{\omega} \mathcal{\widehat{J}}_{\rm LL}(\omega) = \nabla_{\omega} D_{\textcolor{red}{r(v)}}(q(v)q_{\omega_1}(h|v)\Vert p_{\omega_2}(v,h)) \\
        + \nabla_{\omega_2} \mathbb{E}_{p_{\omega_2}(v,h)}[\mathcal{E}_{\psi^{\prime}}(v)] + \textcolor{blue}{\nabla_{\omega_2} D_{\rm KL}(p_{\omega_2}(v,h)\Vert q_{\omega_1}(h|v))},
    \end{gathered}
    \label{eq:pllgr}
\end{equation}
\begin{equation}
\setlength{\abovedisplayskip}{5pt}
\setlength{\belowdisplayskip}{5pt}
    \begin{gathered}
        \nabla_{\psi^{\prime}} \mathcal{\widehat{J}}_{\rm UL}(\psi^{\prime}) = \mathbb{E}_{q(v)}[\nabla_{\psi^{\prime}}\mathcal{E}_{\psi^{\prime}}(v)] - \mathbb{E}_{p_{\omega_2}(v,h)}[\nabla_{\psi^{\prime}}\mathcal{E}_{\psi^{\prime}}(v)]. 
    \end{gathered}
    \label{eq:pulgr}
\end{equation}
The offset gradients are stabler than the original one, as the opposite terms are integrated into one level, however weakens the adversarial learning as we explained in \cref{sec:analy}. 
\begin{algorithm}
\caption{Bi-level doubly variational learning for EBLVM by alternative stochastic gradient descent}
\label{alg:bidvl}
\hspace*{0.02in}{\bf Input:}
Learning rate schemes $\alpha$ and $\beta$; randomly initialized network parameters $\omega$ and $\psi^{\prime}$; number of lower-level step $N$
\begin{algorithmic}[1]
\REPEAT
    \STATE Sample a batch of data
    \FOR{$n=1,\cdots,N$}
        \STATE Optimize the lower-level subproblem (\ref{eq:newll}) by \cref{eq:nllgr} (or \cref{eq:pllgr}): \\
        $\omega \leftarrow \omega - \alpha\nabla_{\omega} \mathcal{J}_{\rm LL}(\omega)$
    \ENDFOR
    \STATE Optional \textit{gradient unrolling} following \cite{Metz2017UnrolledGA}
    \STATE Optimize the upper-level subproblem (\ref{eq:newul}) by \cref{eq:nulgr} (or \cref{eq:pulgr}): \\
    $\psi^{\prime} \leftarrow \psi^{\prime} - \beta\nabla_{\psi^{\prime}} \mathcal{J}_{\rm UL}(\psi^{\prime},\omega_1)$ \\ 
    $\omega_1 \leftarrow \omega_1 - \alpha\nabla_{\omega_1} \mathcal{J}_{\rm UL}(\psi^{\prime},\omega_1)$
\UNTIL{Convergence or reaching certain threshold}
\end{algorithmic}
\end{algorithm}

\subsection{Algorithm implementation}
\label{sec:algim}

The training procedure is summarized in \cref{alg:bidvl}, and in this section, a specific implementation is introduced. Our model consists of an EBM, an inference model and a generative model, corresponding to the marginal energy-based distribution $p_{\psi^{\prime}}(v)$, the variational posterior $q_{\omega_1}(h|v)$ and the variational joint distribution $p_{\omega_2}(v,h)$, respectively. 

For the generative model, BiDVL has many choices, \eg implicit generative models (IGMs) or flows \cite{JimenezRezende2015VariationalIW}. In this work, we simply utilize an IGM, from which a sample is obtained via passing a latent variable $h$ through a deterministic generator, \ie, $v=G_{\omega_2}(h)$, where $h$ is sampled from a known distribution $p(h)$. For the inference model, we employ the standard Gaussian encoder of VAE, ${\mathcal{N}(\mu_{\omega_1}(v),\Sigma_{\omega_1}(v))}$. Sampling from the inference model typically adopts the reparametrization trick, \ie $h=\mu_{\omega_1}(v)+\Sigma_{\omega_1}(v)\cdot\epsilon$, where $\epsilon$ denotes the Gaussian noise. 

\subsection{Discussion}
\label{sec:discuss}

In this section, we present some additional properties of our BiDVL under the specific implementation. By adopting the structure of VAE, \cref{eq:nllgr} contains an importance weighted VAE loss, where we practically use the importance weighted reconstruction loss and KL regularization in \cref{eq:nllgr}. Meanwhile, $D_{\rm KL}(p_{\omega_2}(v,h)\Vert q_{\omega_1}(h|v))$, which forms another reconstruction over latent space, takes part in a cycle reconstruction, linking the models tightly. 

Besides, there are two chase games in \cref{eq:newll,eq:newul,eq:nllgr,eq:nulgr}:

1) The first chase game is demonstrated between the marginal EBM $p_{\psi^{\prime}}$ and the IGM $p_{\omega_2}$. In lower level, $p_{\omega_2}$ tends to chase $p_{\psi^{\prime}}$, while in upper level, $p_{\psi^{\prime}}$ turns to data distribution and escapes from $p_{\omega_2}$. The game formulates an adversarial learning similar to GAN, however discriminator guided IGM typically has limited capacity for assigning density on the learned support \cite{Arbel2021GeneralizedEB}, while in BiDVL, IGM is guided by a sophisticated energy-based probability model. 

2) Opposite terms in \cref{eq:nllgr,eq:nulgr} contribute to another chase game between variational models. In lower level, $q_{\omega_1}$ and $p_{\omega_2}$ are drew close over latent space, while in upper level, $q_{\omega_1}$ attempts to escape from the other. Optimizing $\omega_1$ to increase $D_{\rm KL}(p_{\omega_2}(v,h)\Vert q_{\omega_1}(h|v))$ in the upper level changes the mean and variance of inference model aimlessly, resulting in unstable training. Inspired by the adversarial nature between the reconstruction part and the KL regularization part in VAE loss, we instead minimize $D_{\rm KL}(q_{\omega_1}(h|v)||p(h)), v\sim{p_{\omega_2}(v)}$ in the upper level to interpret the chase game. In fact, it provides $q_{\omega_1}$ a fixed target for stable training. 

All of the properties such as importance weighted loss, the additional chase game and the bi-level optimization scheme are originated from the unified framework, implying the consistency for learning EBLVMs. 

\section{Experiments}

We evaluate BiDVL on three tasks: image generation, test image reconstruction and out-of-distribution (OOD) detection. Test image reconstruction are conducted to evaluate the ability of modelling latent space, the rest experiments are designed mainly following \cite{Han2020JointTO}. In the main paper, experiments are demonstrated on three commonly adopted benchmark datasets including CIFAR-10 \cite{Krizhevsky2009LearningML}, SVHN \cite{Netzer2011ReadingDI} and CelebA \cite{Liu2015DeepLF}. For CIFAR-10 and SVHN, images are resized to $32\times32$. For CelebA, we resize images to $32\times32$ and $64\times64$ to construct CelebA-32 and CelebA-64. All datasets are scaled to $[-1,1]$ for pre-processing. 

Our model consists of an EBM, an inference model and an IGM. The structure of variational models is following \cite{Han2020JointTO}, which simply cascades several convolutional layers. For stable training, batch normalization \cite{Ioffe2015BatchNA} layers are plugged between convolutional layers. The EBM uses convolutional layers to map a sample to real-valued energy with spectral normalization to ensure the constraint on $\nabla_{\psi}\mathcal{E}_{\psi}(v,h)$ discussed in \cref{sec:diver}. Parameters are optimized via Adam \cite{Kingma2015AdamAM}. We set the number of lower-level steps $N$ to one and ignore the \textit{gradient unrolling} for faster and stabler training in our experiments. Some major choices are discussed in \cref{sec:analy}. 

In order to evaluate on larger scale images, we also conduct experiments on CelebA resized to $128\times128$ with stronger Resnet-based structure in \cref{sec:addex}. 

\subsection{Image generation}
\label{sec:imgen}

\begin{figure*}
  \centering
  \includegraphics[width=0.3\linewidth]{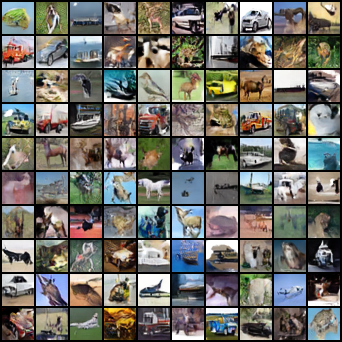}
  \includegraphics[width=0.3\linewidth]{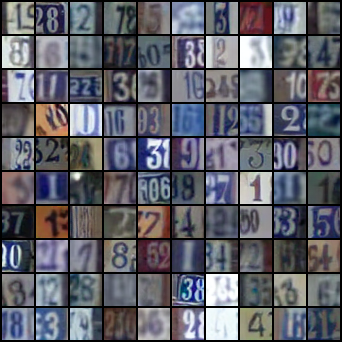}
  \includegraphics[width=0.3\linewidth]{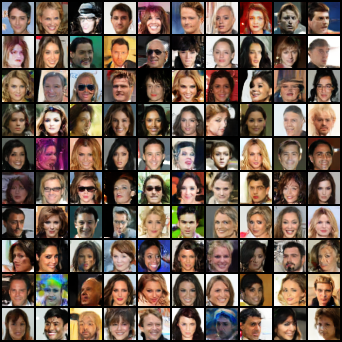}
  \caption{Randomly generated images. Left: CIFAR-10. Middle: SVHN. Right: CelebA-32.}
   \label{fig:1}
\end{figure*}

\begin{figure}[t]
  \centering
  \includegraphics[width=0.8\linewidth]{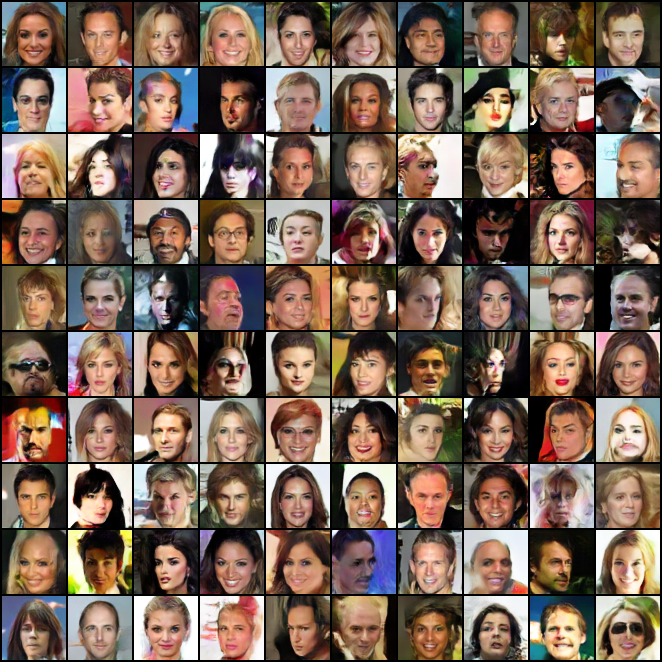}
   \caption{Randomly generated images on CelebA-64.}
   \label{fig:2}
\end{figure}

We show the proposed model can generate realistic images with visual similarities to the training images by conducting experiments on CIFAR-10, SVHN and CelebA. \Cref{fig:1} shows the result of $32\times32$ images randomly generated by proposed model and \cref{fig:2} shows the result of $64\times64$ images on CelebA. We find the generated images maintain diversity, which confirms that the learned EBLVM well covers the most modes of data. 

We adopt Frechet Inception Distance (FID) \cite{Heusel2017GANsTB} to reflect the sample quality and the baseline models are chosen from EBMs, VAEs and GANs to align with the perspectives in \cref{sec:discuss}. Divergence triangle \cite{Han2020JointTO} is an energy-guided VAE with a similar model structure to us. IGEBM \cite{Du2019ImplicitGA} and GEBM \cite{Arbel2021GeneralizedEB} are MCMC-based EBMs while VERA \cite{Grathwohl2021NoMF} is MCMC-free. SNGAN \cite{Miyato2018SpectralNF} adopts the structure of Resnet and gets 21.7 FID on CIFAR-10. The recent exponential tilting models which formulate a two-stage refinement procedure with inefficient MCMC sampling, are not considered here. In \cref{tab:imge1}, we evaluate our best model on CIFAR-10 and CelebA-64. \Cref{tab:imge2} reports the FID on CelebA-32. Our model achieves superior performance than baseline models, especially getting 20.75 FID on CIFAR-10, even without elaborate structure such as Resnet. 
\begin{table}
  \centering
  \begin{tabular}{lcc}
    \toprule
    Model & CIFAR-10 & CelebA-64 \\
    \midrule
    VAE \cite{Kingma2014Autoencoding} & 109.5 & 99.09 \\
    Divergence triangle \cite{Han2020JointTO} & 30.1 & 24.7 \\
    IGEBM \cite{Du2019ImplicitGA} & 40.58 & - \\
    GEBM \cite{Arbel2021GeneralizedEB} & 23.02 & - \\
    VERA \cite{Grathwohl2021NoMF} & 27.5 & - \\
    SNGAN \cite{Miyato2018SpectralNF} & 21.7 & 50.4 \\
    BiDVL (ours) & \textbf{20.75} & \textbf{17.24} \\
    \bottomrule
  \end{tabular}
  \caption{Evaluation of sample quality on CIFAR-10 and CelebA-64 via FID.}
  \label{tab:imge1}
\end{table}
\begin{table}
  \centering
  \begin{tabular}{lc}
    \toprule
    Model & CelebA-32 \\
    \midrule
    VAE \cite{Kingma2014Autoencoding} & 38.76 \\
    DCGAN \cite{Radford2016UnsupervisedRL} & 12.50 \\
    FCE \cite{Gao2020FlowCE} & 12.21 \\
    GEBM \cite{Arbel2021GeneralizedEB} & 5.21 \\
    BiDVL (ours) & \textbf{4.47} \\
    \bottomrule
  \end{tabular}
  \caption{Evaluation of sample quality on CelebA-32 using FID.}
  \label{tab:imge2}
\end{table}

\subsection{Image reconstruction}
\label{sec:imrec}

\begin{figure}[t]
  \centering
  \includegraphics[width=0.48\linewidth]{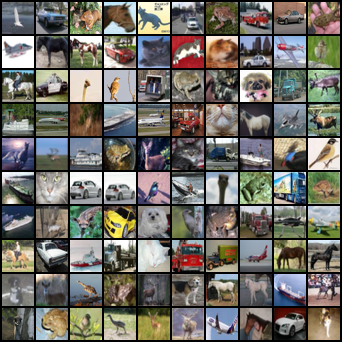}
  \includegraphics[width=0.48\linewidth]{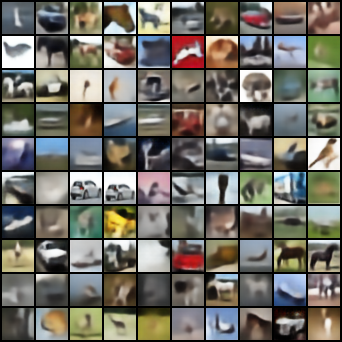}
  \includegraphics[width=0.48\linewidth]{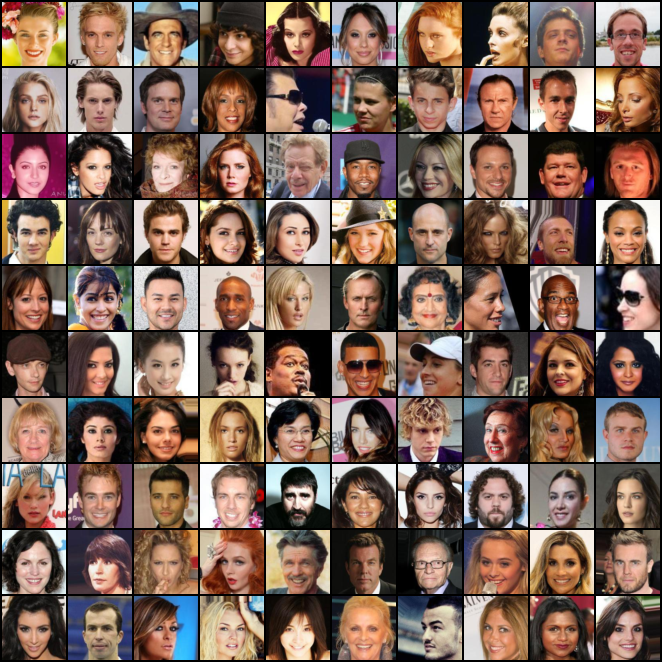}
  \includegraphics[width=0.48\linewidth]{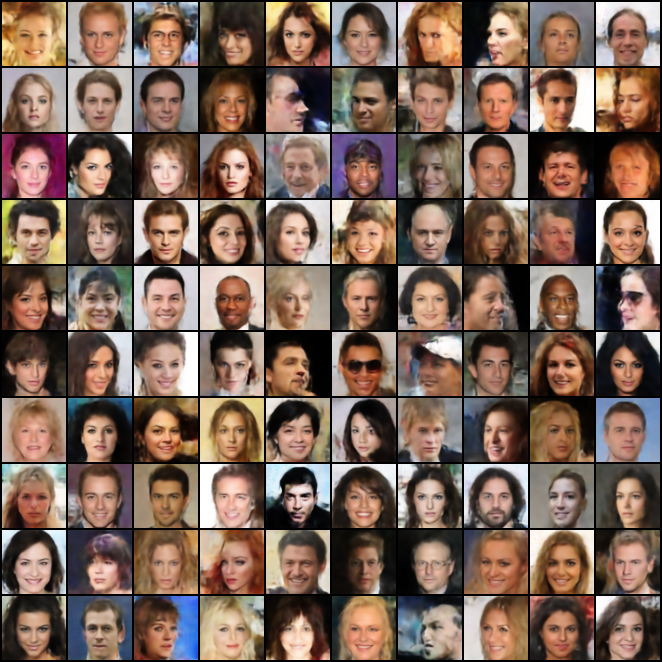}
   \caption{Test image reconstruction. Top: CIFAR-10. Bottom: CelebA-64. Left: test images. Right: reconstruction images.}
   \label{fig:3}
\end{figure}

In this section, we show our model can learn the low-dimensional structure of training samples by assessing the performance on testing image reconstruction. We must note that, EBMs and GANs are incapable of image reconstruction without an inference model, while an inference model is adopted as the variational posterior in our propose model, helping to accomplish the infer-reconstruct procedure. 
Experimentally, our model achieves a low reconstruction error on both CIFAR-10 and CelebA-64 test datasets, because the reconstruction behavior over latent variables helps to further enhance the alignment between visible space and latent space. We compare our model, which has the best generation in \cref{sec:imgen}, to baseline models in \cref{tab:imre}, with rooted mean square error (RMSE) as the metric. Reconstruction images shown in \cref{fig:3} demonstrate our model can capture the major information in testing images, and the low reconstruction error implies that the information of real data flows to the EBLVM through IGM effectively. 
\begin{table}
  \centering
  \begin{tabular}{lcc}
    \toprule
    Model & CIFAR-10 & CelebA-64 \\
    \midrule
    VAE \cite{Kingma2014Autoencoding} & 0.192 & 0.197 \\
    ALICE \cite{Li2017ALICETU} & 0.185 & 0.214 \\
    SVAE \cite{Pu2018SymmetricVA} & 0.258 & 0.209 \\
    Divergence triangle \cite{Han2020JointTO} & 0.177 & 0.190 \\
    BiDVL (ours) & \textbf{0.168} & \textbf{0.187} \\
    \bottomrule
  \end{tabular}
  \caption{Evaluation of testing image reconstruction on CIFAR-10 and CelebA-64 using RMSE.}
  \label{tab:imre}
\end{table}

\subsection{Out-of-distribution detection}
\label{sec:ood}

The EBLVM defines an unnormalized density function which can be used for detecting OOD samples. Generative models like GANs are infeasible for OOD because the model distribution is implicitly defined, while likelihood models like VAEs and Flows are accused of overestimating OOD regions, and as a result, fails to distinguish OOD samples. However the log unnormalized marginal density of EBLVM, \ie, the negative marginal energy $-\mathcal{E}(v)$, is trained to assign low value on OOD regions and high value on data regions, which is suitable for our case. Since the marginal part of the decoupled EBLVM can model the visible space faithfully, we regard $-\mathcal{E}_{\psi^{\prime}}(v)$ as the critic of samples. We mainly consider three OOD datasets: uniform noise, SVHN and CelebA. Our proposed model is trained on CIFAR-10 whose testing part is in-distribution dataset. Following \cite{Du2019ImplicitGA}, area under the ROC curve (AUROC) is used as our evaluation metric. Since training EBM is of a high variance process found in related work \cite{Han2020JointTO}, our best model is chosen from the early stage of training. \Cref{tab:ood} shows our model achieves compatible performance with most of baseline chosen from recent EBMs, VAEs and flows. Surprisingly, we find our proposed model outperforms JEM slightly, though JEM is an energy-based classifier model supposed to be good at OOD detection. Moreover, VERA \cite{Grathwohl2021NoMF}, a recent MCMC-free method for learning EBMs, performs quite well on SVHN dataset, while our model is better on CelebA. 
\begin{table}
  \centering
  \begin{tabular}{lccc}
    \toprule
    Model & Random & SVHN & CelebA \\
    \midrule
    IGEBM \cite{Du2019ImplicitGA} & 1.0 & 0.63 & 0.7 \\
    Glow \cite{Kingma2018GlowGF} & 1.0 & 0.24 & 0.57 \\
    SVAE \cite{Pu2018SymmetricVA} & 0.29 & 0.42 & 0.52 \\
    Divergence triangle \cite{Han2020JointTO} & 1.0 & 0.68 & 0.56 \\
    JEM \cite{Grathwohl2020YourCI} & 1.0 & 0.67 & 0.75 \\
    VERA \cite{Grathwohl2021NoMF} & 1.0 & \textbf{0.83} & 0.33 \\
    BiDVL (ours) & 1.0 & 0.76 & \textbf{0.77} \\
    \bottomrule
  \end{tabular}
  \caption{Out-of-distribution detection on uniform, SVHN, CelebA test datasets, with CIFAR-10 as the in-distribution dataset. We report the AUROC of negative free energy.}
  \label{tab:ood}
\end{table}

\subsection{Ablations and analysis}
\label{sec:analy}

We first investigate how the modifications proposed in \cref{sec:cobidvl} solve the problems by enhancing the learning stability. We conduct experiments on the original BiDVL with canonical KL, but it is unstable (even diverse) on CIFAR. With a similar phenomenon, only decoupling the EBLVM is still hard to converge. On the other side, the training of undecoupled EBLVM improves a lot with the symmetric KL and gets 32.10 FID on CIFAR, nonetheless much worse than our complete model that gets 20.75 shown in \cref{sec:imgen}. 

Next we study the rest choices in BiDVL. Since computing the importance ratio $r(v)$ is nontrivial, we heuristically estimate it with a basic term and a bias term.
\begin{equation}
\setlength{\abovedisplayskip}{5pt}
\setlength{\belowdisplayskip}{5pt}
    \begin{split}
        r(v) =&\ \frac{\mathbb{E}_{p_{\psi^{\prime}}(v)}[q(v)]}{q(v)} \frac{\exp{(-\mathcal{E}_{\psi^{\prime}}(v))}}{\mathbb{E}_{q(v)}[\exp{(-\mathcal{E}_{\psi^{\prime}}(v))}]} \\
        \approx&\ r^{\prime} \frac{\exp{(-\mathcal{E}_{\psi^{\prime}}(v))}}{\mathbb{E}_{q(v)}[\exp{(-\mathcal{E}_{\psi^{\prime}}(v))}]},
    \end{split}
    \label{eq:is}
\end{equation}
where the basic term $r^{\prime}$ corresponds to the average ratio, the bias term means samples with higher density under model distribution should be learned more. As training EBM is a high-variance process, which makes the estimate of the bias ratio noisy, we use $\mathrm{Sigmoid}(\mathbb{E}_{q(v)}[\mathcal{E}_{\psi^{\prime}}(v)]-\mathcal{E}_{\psi^{\prime}}(v))$ to replace the bias term but still leads to slightly inferior performance, consequently we ignore the bias term. 

Furthermore, we found the proposed model performs quite sensitive to the basic term, as shown in \cref{tab:basra} studying the influences of varying basic ratio on CIFAR. 
\begin{table}
  \centering
  \begin{tabular}{lccccc}
    \toprule
    $r^{\prime}$ & 0.01 & 0.05 & 0.1 & 0.5 & 1.0 \\
    \midrule
    FID & 22.62 & 20.75 & 21.90 & 26.82 & 29.72 \\
    \bottomrule
  \end{tabular}
  \caption{The influences of varying basic ratio on FID. Models are trained on CIFAR-10.}
  \label{tab:basra}
\end{table}
The generation quality degrades as the basic ratio increases and perform pretty worse with $r^{\prime}=1.0$. It corresponds to the intuition that the importance ratio is supposed to be small on average since the energy-based distribution is mild and typically assigns density on unreal samples. Finally, we use $0.05$ on $32\times32$ datasets and $0.1$ on $64\times64$ datasets. 

Then we study the effect of offsetting the opposite terms in \cref{eq:nllgr,eq:nulgr}. The implementation details of not offset algorithm are follow \cref{sec:discuss}. For comparison, we evaluate the algorithm without all opposite terms concerned. Not offset algorithm performs unstable on CelebA-64, even not converge sometimes. \Cref{tab:offset} shows the influence. 
\begin{table}
  \centering
  \begin{tabular}{lccc}
    \toprule
     & w/o & offset & not offset \\
    \midrule
    FID & 20.24 & 20.75 & 21.54 \\ 
    RMSE & 0.175 & 0.168 & 0.170 \\
    \bottomrule
  \end{tabular}
  \caption{The influences of offsetting on FID and RMSE. Models are trained on CIFAR-10.}
  \label{tab:offset}
\end{table}
We find the $\mathrm{Sigmoid}$ estimation of bias term helps the not offset algorithm stabilizing, however it tends to overfit early and performs slightly worse. Not offset version achieves $0.81$ AUROC on SVHN and $0.75$ AUROC on CelebA, implying the adversarial learning may help BiDVL better modelling. Algorithm without terms concerned is slightly better on generation, but is inferior on reconstruction and OOD detection. It implies the reconstruction over latent variables enhances the alignment. Our formal experiments are conducted by the offset version. 

\section{Conclusion}
\label{sec:concl}

In this paper, we introduce BiDVL, a MCMC-free framework to efficiently learn EBLVMs. For image task, we achieve a decoupled EBLVM, consisting of a marginal energy-based distribution and a structural posterior, and then choose a symmetric KL divergence in the lower level. Experiments show the impressive generative power on images concerned. In this work, simply stacking several convolutional layers largely restricts our model from scaling to a large image dataset. In future work, the deeper and more sophisticated designed networks shall be adopted to evaluate BiDVL on larger-scale images. 

\noindent \textbf{Broader impacts.} The method is for training EBLVMs, and the trained models have the ability of generating fake content, which may be used with potential malicious. 

\noindent \textbf{Acknowledgement.} This work is partially supported by the National Natural Science Foundation of China (62141604, 61972016, 62032016, 62106012, 62076016, 62101245), Natural Science Foundation of Beijing Municipality (L191007).

{\small
\bibliographystyle{ieee_fullname}\bibliography{BiDVL}
}

\clearpage

\appendix

\section{Bi-level optimization}

\subsection{Metrics for lower-level subproblem} 
\label{sec:diver}

\textbf{IPMs.} We start from the gradient of standard objective for EBLVM \ie\cref{eq:split}:
\begin{gather}
    \omega^*(\psi) = \arg\min_{\omega\in\Omega} \notag \\
    \left\Vert \mathbb{E}_{p(v)p_{\psi}(h|v)} \left[\nabla_{\psi}\mathcal{E}_{\psi} \right] - \mathbb{E}_{p(v)q_{\omega_1}(z|a)} \left[\nabla_{\psi}\mathcal{E}_{\psi} \right] \right\Vert_{\infty}
    \label{eq:a1}\\
    + \left\Vert \mathbb{E}_{p_{\omega_2}(v,h)} \left[\nabla_{\psi}\mathcal{E}_{\psi} \right] - \mathbb{E}_{p_{\psi}(v,h)} \left[\nabla_{\psi}\mathcal{E}_{\psi} \right] \right\Vert_{\infty}.
    \label{eq:a2}
\end{gather}
Then we bound two infinite norms from above by general integral probability metrics (IPMs), we only show derivation about term (\ref{eq:a2}) and the derivation about term (\ref{eq:a1}) can be obtained by the same way:
\begin{equation}
    \begin{split}
        &\ \left\Vert \mathbb{E}_{p_{\omega_2}(v,h)} \left[\nabla_{\psi}\mathcal{E}_{\psi} \right] - \mathbb{E}_{p_{\psi}(v,h)} \left[\nabla_{\psi}\mathcal{E}_{\psi} \right] \right\Vert_{\infty} \\
        \leq&\ \sup_{f:\mathcal{V}\times\mathcal{H}\rightarrow\mathbb{R},f\in\mathcal{F}} \left| \mathbb{E}_{p_{\omega_2}(v,h)} \left[f(v,h) \right] - \mathbb{E}_{p_{\psi}(v,h)} \left[f(v,h) \right] \right| \\
        =&\ D_{\mathcal{F}}(p_{\omega_2}(v,h),p_{\psi}(v,h)),
    \end{split} \label{eq:a3}
\end{equation}
where $\mathcal{F}$ is a class of scalar function over space $\mathcal{V}\times\mathcal{H}$, $D_{\mathcal{F}}$ denotes the IPM induced by $\mathcal{F}$. Notice $\mathcal{F}$ depends on the gradient of energy function $\nabla_{\psi}\mathcal{E}_{\psi}(v,h)$, we thus propose some assumptions about $\nabla_{\psi}\mathcal{E}_{\psi}(v,h)$ for special cases:

(1) Assume the infinite norm of each component, \ie $\Vert\nabla_{\psi}\mathcal{E}_{\psi}(v,h)_{i}\Vert_{\infty}$, is bounded by a constant $C$, then
\begin{equation*}
    (\ref{eq:a2}) \leq C\cdot D_{\rm TV}(p_{\omega_2}(v,h),p_{\psi}(v,h)),
\end{equation*}
where $D_{\rm TV}$ denotes the total variation distance corresponding to $\mathcal{F}=\{f:\Vert f \Vert_{\infty}\le 1 \}$.

(2) Assume each component $\nabla_{\psi}\mathcal{E}_{\psi}(v,h)_{i}$ is $C$-Lipschitz, then
\begin{equation*}
    (\ref{eq:a2}) \leq C\cdot W_{1}(p_{\omega_2}(v,h),p_{\psi}(v,h)),
\end{equation*}
where $W_{1}$ denotes the 1-Wasserstein distance corresponding to $\mathcal{F}=\{f:f\text{ is 1-Lipschitz} \}$.

(3) Assume the norm of each component defined in reproducing kernel Hilbert space $\mathcal{H}$, i.e. $\Vert\nabla_{\psi}\mathcal{E}_{\psi}(v,h)_{i}\Vert_{\mathcal{H}}$, is bounded by a constant $C$, then
\begin{equation*}
    (\ref{eq:a2}) \leq C\cdot {\rm MMD}(p_{\omega_2}(v,h),p_{\psi}(v,h)),
\end{equation*}
where ${\rm MMD}$ denotes the maximum mean discrepancy corresponding to $\mathcal{F}=\{f:\Vert f \Vert_{\mathcal{H}}\le 1 \}$.

\textbf{Practical choices.} Assumption (2) is hard to verify because both $p_{\psi}(h|v)$ and $p_{\psi}(v,h)$ are intractable in our setting, so that we can not directly use the metric for optimization. In practice, we resort to moderate metrics under mild assumptions:

1. Notice that assumption (1) is typically mild in practice and we can consider the generally adopted KL divergence by Pinsker's inequality
\begin{equation*}
    2D_{\rm TV}(p_{w_2},p_{\psi})^2 \leq D_{\rm KL}(p_{w_2}\Vert p_{\psi}),
\end{equation*}
where the equality holds only if $p_{w_2}$ equals $p_{\psi}$. KL divergence has a stronger convergence than many other divergence metrics and we show that it is feasible in our work. 

2. Under assumption (3), ${\rm MMD}^2(p,q)=\mathbb{E}[k(x,x^{\prime})-2k(x,y)+k(y,y^{\prime})]$, where $x,x^{\prime}$ and $y,y^{\prime}$ are \iid drew from $p$ and $q$, respectively. However, it is impossible to obtain its gradient \wrt $w$ by Monte Carlo estimation. The kernelized Stein discrepancy (KSD) is a special MMD with a kernel $u_{q}(x,x^{\prime})$ depending on $q$:
\begin{gather*}
    {\rm KSD}(p,q) = \mathbb{E}_{x,x^{\prime}\sim p}[u_q(x,x^{\prime})] \\
    u_q(x,x^{\prime}) = \mathbf{s}_q(x)^{\top}k(x,x^{\prime})\mathbf{s}_q(x^{\prime})+\mathbf{s}_q(x)^{\top}\nabla_{x^{\prime}}k(x,x^{\prime}) \\
    +\nabla_{x}k(x,x^{\prime})^{\top}\mathbf{s}_q(x^{\prime})+\mathop{\rm trace}(\nabla_{x,x^{\prime}}k(x,x^{\prime})),
\end{gather*}
where $\mathbf{s}_q(x)=\nabla_{x}\log q(x)$ is known as the score function. Score function is a gradient \wrt $x$, so it eliminates the intractable partition function which is independent of variable $x$. We also refer to Fisher divergence as a moderate metric in our setting which is further stronger than KL, total variation and KSD. Using Fisher divergence in fact corresponds to score matching widely used in learning generative models. 

\subsection{Derivations}
\label{sec:derive}

In this subsection, we provide some supplemental derivations. We start from the \cref{eq:part} with focus on the second part in \cref{eq:part} and the first part can be derived in a same way: 
\begin{equation*}
    \begin{gathered}
        \frac{\partial D_{\rm KL}(p_{\omega_2(\psi)}(v,h)\Vert p_{\psi}(v,h))}{\partial \psi} \\
        = \frac{\partial D_{\rm KL}(p_{\omega_2}(v,h)\Vert p_{\psi}(v,h))}{\partial \psi}|_{\omega_2=\omega_2(\psi)} \\
        + \left( \frac{\partial\omega_2(\psi)}{\partial\psi^{\top}} \right)^{\top} \frac{\partial D_{\rm KL}(p_{\omega_2}(v,h)\Vert p_{\psi}(v,h))}{\partial \omega_2}|_{\omega_2=\omega_2(\psi)}, 
    \end{gathered}
\end{equation*}
where $\frac{\partial\omega_2(\psi)}{\partial\psi}$ is the Jacobian. We then look into its first term as follow where we use $\nabla_{\psi}$ for clarity:  
\begin{equation*}
    \begin{split}
    &\ \nabla_{\psi}D_{\rm KL}(p_{\omega_2}(v,h)\Vert p_{\psi}(v,h))|_{\omega_2=\omega_2(\psi)} \\
    =&\ \nabla_{\psi} \left[ \int p_{\omega_2}(v,h)\log\frac{p_{\omega_2}(v,h)}{p_{\psi}(v,h)} \mathrm{d}h\mathrm{d}v \right]_{\omega_2=\omega_2(\psi)} \\
    =&\ \nabla_{\psi} \left[ -\int p_{\omega_2}(v,h)\log\frac{\exp{(-\mathcal{E}_{\psi}})}{\mathcal{Z}(\psi)} \mathrm{d}h\mathrm{d}v \right]_{\omega_2=\omega_2(\psi)} \\
    =&\ \mathbb{E}_{p_{\omega_2(\psi)}(v,h)} \left[\nabla_{\psi}\mathcal{E}_{\psi} \right] + \nabla_{\psi}\log\mathcal{Z}(\psi), 
    \end{split}
\end{equation*}
where the partition function $\mathcal{Z}(\psi)$ is independent of variables $v$ and $h$. At last, we obtain \cref{eq:part} by 
\begin{equation*}
    \begin{gathered}
        \frac{\partial D_{\rm KL}(q(v)q_{\omega_1}(h|v)\Vert p_{\psi}(v,h))}{\partial \psi}|_{\omega_1=\omega_1(\psi)} \\
        - \frac{\partial D_{\rm KL}(p_{\omega_2}(v,h)\Vert p_{\psi}(v,h))}{\partial \psi}|_{\omega_2=\omega_2(\psi)} \\
        = \mathbb{E}_{q(v)q_{\omega_1(\psi)}(h|v)} \left[\nabla_{\psi}\mathcal{E}_{\psi} \right] + \nabla_{\psi}\log\mathcal{Z}(\psi) \\
        - \mathbb{E}_{p_{\omega_2(\psi)}(v,h)} \left[\nabla_{\psi}\mathcal{E}_{\psi} \right] - \nabla_{\psi}\log\mathcal{Z}(\psi) \\
        = \mathbb{E}_{q(v)q_{\omega_1(\psi)}(h|v)} \left[\nabla_{\psi}\mathcal{E}_{\psi} \right] - \mathbb{E}_{p_{\omega_2(\psi)}(v,h)} \left[\nabla_{\psi}\mathcal{E}_{\psi} \right]. 
    \end{gathered}
\end{equation*}

Furthermore, recall \cref{eq:ulgra}, we have: 
\begin{equation}
\begin{aligned}
        &\ \frac{\partial\mathcal{J}_{\rm UL}(\psi,\omega)}{\partial\psi}|_{\omega=\omega^*(\psi)} = \\
        &\ \mathbb{E}_{q(v)q_{\omega_1^*(\psi)}(h|v)} \left[\nabla_{\psi}\mathcal{E}_{\psi} \right] - \mathbb{E}_{p_{\omega_2^*(\psi)}(v,h)} \left[\nabla_{\psi}\mathcal{E}_{\psi} \right]
\end{aligned}
\label{eq:a4}
\end{equation}
\begin{equation}
\begin{gathered}
    \left( \frac{\partial\omega^*(\psi)}{\partial\psi^{\top}} \right)^{\top}\frac{\partial\mathcal{J}_{\rm UL}(\psi,\omega)}{\partial\omega}|_{\omega=\omega^*(\psi)} = \\
    \left( \frac{\partial\omega_1^*(\psi)}{\partial\psi^{\top}} \right)^{\top} \frac{\partial D_{\rm KL}(q(v)q_{\omega_1}(h|v)\Vert p_{\psi}(v,h))}{\partial \omega_1}|_{\omega_1=\omega_1^*(\psi)} \\
    - \left( \frac{\partial\omega_2^*(\psi)}{\partial\psi^{\top}} \right)^{\top} \frac{\partial D_{\rm KL}(p_{\omega_2}(v,h)\Vert p_{\psi}(v,h))}{\partial \omega_2}|_{\omega_2=\omega_2^*(\psi)}, \label{eq:a5}
\end{gathered}
\end{equation}

\subsection{Proof and properties}
\label{sec:proof}

We next proof the equivalence of BLO problem (\ref{eq:ll},\ref{eq:ul}) and the original one, under the unparametric assumption.
\begin{proof}[\textbf{Proof of Theorem  \ref{thm1}}]
Suppose for $\psi\in\Psi$ we have $\hat{\omega}\in\Omega$ such that $D(q(v)q_{\hat{\omega}_1}(h|v),q(v)p_{\psi}(h|v))=0$ and $D(p_{\hat{\omega}_2}(v,h),p_{\psi}(v,h))=0$, then $0\leq\mathcal{J}_{\rm LL}(\psi,\omega^*(\psi))\leq\mathcal{J}_{\rm LL}(\psi,\hat{\omega})=0$, thus $\mathcal{J}_{\rm LL}(\psi,\omega^*(\psi))=0$. In other words, $\omega^*(\psi)$ satisfies $q_{\omega_1^*(\psi)}(h|v)=p_{\psi}(h|v),p_{\omega_2^*(\psi)}(v,h)=p_{\psi}(v,h)$. Then we have
\begin{equation}
    \begin{gathered}
        |\mathcal{J}_{\rm UL}(\psi,\omega^*(\psi)) - \mathcal{J}(\psi) | \\
        = | D_{\rm KL}(q(v)q_{\omega_1^*(\psi)}(h|v)\Vert p_{\psi}(v,h)) \\
        -\ D_{\rm KL}(p_{\omega_2^*(\psi)}(v,h)\Vert p_{\psi}(v,h)) - D_{\rm KL}(q(v)\Vert p_{\psi}(v)) | \\
        = \bigg| \mathbb{E}_{q(v)q_{\omega_1^*(\psi)}(h|v)} \left[\log\frac{q(v)q_{\omega_1^*(\psi)}(h|v)}{p_{\psi}(v)p_{\psi}(h|v)} \right] \\
        -\ D_{\rm KL}(p_{\omega_2^*(\psi)}(v,h)\Vert p_{\psi}(v,h)) - \mathbb{E}_{q(v)} \left[\log\frac{q(v)}{p_{\psi}(v)}  \right] \bigg| \\
        = \bigg| \mathbb{E}_{q(v)q_{\omega_1^*(\psi)}(h|v)} \left[\log\frac{q_{\omega_1^*(\psi)}(h|v)}{p_{\psi}(h|v)} \right] \\
        -\ D_{\rm KL}(p_{\omega_2^*(\psi)}(v,h)\Vert p_{\psi}(v,h)) \bigg| \\
        = | D_{\rm KL}(q(v)q_{\omega_1^*(\psi)}(h|v)\Vert q(v)p_{\psi}(h|v)) \\
        -\ D_{\rm KL}(p_{\omega_2^*(\psi)}(v,h)\Vert p_{\psi}(v,h)) | \\
        \leq D_{\rm KL}(q(v)q_{\omega_1^*(\psi)}(h|v)\Vert q(v)p_{\psi}(h|v)) \\
        +\ D_{\rm KL}(p_{\omega_2^*(\psi)}(v,h)\Vert p_{\psi}(v,h)) = 0, 
    \end{gathered}
    \label{eq:a6}
\end{equation}
where the last equation holds because $D_{\rm KL}(P\Vert Q)=0$ is equivalent to $P=Q$. On the other hand, by (\ref{eq:split},\ref{eq:ul},\ref{eq:a4},\ref{eq:a5}), we have 
\begin{equation}
    \begin{gathered}
        \left\Vert \nabla_{\psi}\mathcal{J}(\psi) - \nabla_{\psi}\mathcal{J}_{\rm UL}(\psi,\omega^*(\psi)) \right\Vert_{\infty} \\
        = \bigg\Vert \mathbb{E}_{q(v)p_{\psi}(h|v)} \left[\nabla_{\psi}\mathcal{E}_{\psi} \right] - \mathbb{E}_{q(v)q_{\omega_1^*(\psi)}(h|v)} \left[\nabla_{\psi}\mathcal{E}_{\psi} \right] \\
        +\ \mathbb{E}_{p_{\omega_2^*(\psi)}(v,h)} \left[\nabla_{\psi}\mathcal{E}_{\psi} \right] - \mathbb{E}_{p_{\psi}(v,h)} \left[\nabla_{\psi}\mathcal{E}_{\psi} \right] \\
        -\ \left( \frac{\partial\omega^*(\psi)}{\partial\psi^{\top}} \right)^{\top}\frac{\partial\mathcal{J}_{\rm UL}(\psi,\omega)}{\partial\omega}|_{\omega=\omega^*(\psi)} \bigg\Vert_{\infty} \\
        \leq \Vert \mathbb{E}_{q(v)p_{\psi}(h|v)} \left[\nabla_{\psi}\mathcal{E}_{\psi} \right] -\ \mathbb{E}_{q(v)q_{\omega_1^*(\psi)}(h|v)} \left[\nabla_{\psi}\mathcal{E}_{\psi} \right] \Vert_{\infty} \\
        +\ \Vert \mathbb{E}_{p_{\omega_2^*(\psi)}(v,h)} \left[\nabla_{\psi}\mathcal{E}_{\psi} \right] - \mathbb{E}_{p_{\psi}(v,h)} \left[\nabla_{\psi}\mathcal{E}_{\psi} \right] \Vert_{\infty} \\
        +\ \bigg\Vert \left( \frac{\partial\omega^*(\psi)}{\partial\psi^{\top}} \right)^{\top}\frac{\partial\mathcal{J}_{\rm UL}(\psi,\omega)}{\partial\omega}|_{\omega=\omega^*(\psi)} \bigg\Vert_{\infty}. 
    \end{gathered}
    \label{eq:a7}
\end{equation}
Because $\omega^*(\psi)$ satisfies $q_{\omega_1^*(\psi)}(h|v)=p_{\psi}(h|v)$ and $p_{\omega_2^*(\psi)}(v,h)=p_{\psi}(v,h)$, under nonparametric assumption. Thus we know that $\omega_1=\omega_1^*(\psi),\omega_2=\omega_2^*(\psi)$ are the stationary points of $\min_{\omega_1}D_{\rm KL}(q(v)q_{\omega_1}(h|v)\Vert q(v)p_{\psi}(h|v))$ and $\min_{\omega_2}D_{\rm KL}(p_{\omega_2}(v,h)\Vert p_{\psi}(v,h))$, respectively. Due to 
\begin{equation*}
\begin{split}
      &\ \nabla_{\omega_1}D_{\rm KL}(q(v)q_{\omega_1}(h|v)\Vert q(v)p_{\psi}(h|v)) \\
      =&\ \nabla_{\omega_1}D_{\rm KL}(q(v)q_{\omega_1}(h|v)\Vert p_{\psi}(v,h)),   
\end{split}
\end{equation*}
we have $\frac{\partial\mathcal{J}_{\rm UL}(\psi,\omega)}{\partial\omega}|_{\omega=\omega^*(\psi)}=0$. Consequently we bound \cref{eq:a7} by 
\begin{equation*}
    \begin{gathered}
        \leq C\cdot D(q(v)q_{\omega_1^*(\psi)}(h|v),q(v)p_{\psi}(h|v)) \\
        +\ C\cdot D(p_{\omega_2^*(\psi)}(v,h),p_{\psi}(v,h)) \\
        = C\cdot \mathcal{J}_{\rm LL}(\psi,\omega^*(\psi)) = 0,
    \end{gathered}
\end{equation*}
which is derived from \cref{eq:a3}, $C$ depends on the assumption about the gradient of energy function. At last we have $\nabla_{\psi}\mathcal{J}(\psi)=\nabla_{\psi}\mathcal{J}_{\rm UL}(\psi,\omega^*(\psi))$.
\end{proof}

In fact, the unparametric assumption typically does not hold when we take neural networks as the variational approximators, thus the optima of $\omega$ may not lie in the parameter space $\Omega$. The proof (\ref{eq:a6}) characterizes that the bias of upper-level objective $\mathcal{J}_{\rm UL}(\psi,\omega^*(\psi))$ can be bounded by KL divergences. It means that if we choose KL divergence or certain stronger metric, \eg Fisher divergence, in lower-level subproblem, we can obtain bounds on both sides as $|\mathcal{J}_{\rm UL}(\psi,\omega^*(\psi))-D_{\rm KL}(q(v)\Vert p_{\psi}(v))|\leq C^{\prime}\cdot\min_{\omega\in\Omega} \mathcal{J}_{\rm LL}(\psi,\omega)$, where $C^{\prime}$ depends on the metric we choose and the assumptions about the gradient of energy function $\nabla_{\psi}\mathcal{E}_{\psi}(v,h)$. To ensure the constraint on $\nabla_{\psi}\mathcal{E}_{\psi}(v,h)$, we introduce spectral normalization into the energy network in our experiments. On the other hand, \cref{eq:a7} indicates the upper and lower bound (black curves) in \cref{fig:bidvl}, and the lower-level optimization forces the gradient estimate (the gradient of upper level) to fit the real one.

\section{Additional experiments}
\label{sec:addex}

\begin{figure}[t]
  \centering
  \includegraphics[width=0.96\linewidth]{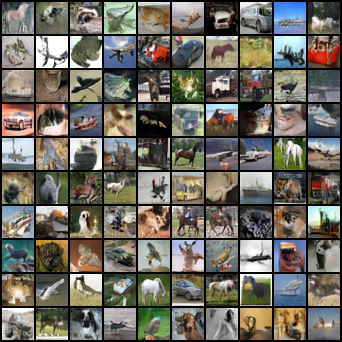}
   \caption{Randomly generated images on CIFAR-10.}
   \label{fig:appcifar}
\end{figure}
\begin{figure}[t]
  \centering
  \includegraphics[width=0.96\linewidth]{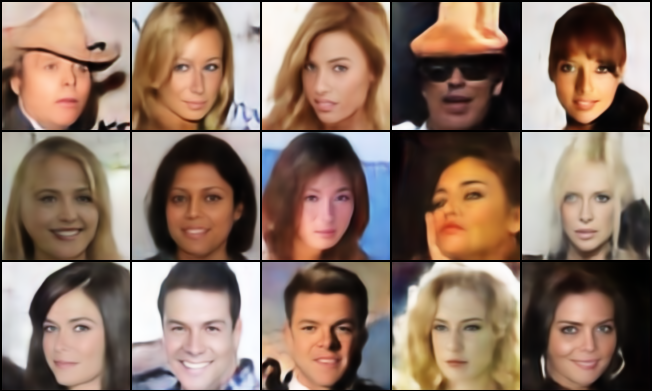}
  \caption{Reconstruction images on CelebA-128.}
  \label{fig:appcelebarecon}
\end{figure}
\begin{figure}[t]
  \centering
  \includegraphics[width=0.96\linewidth]{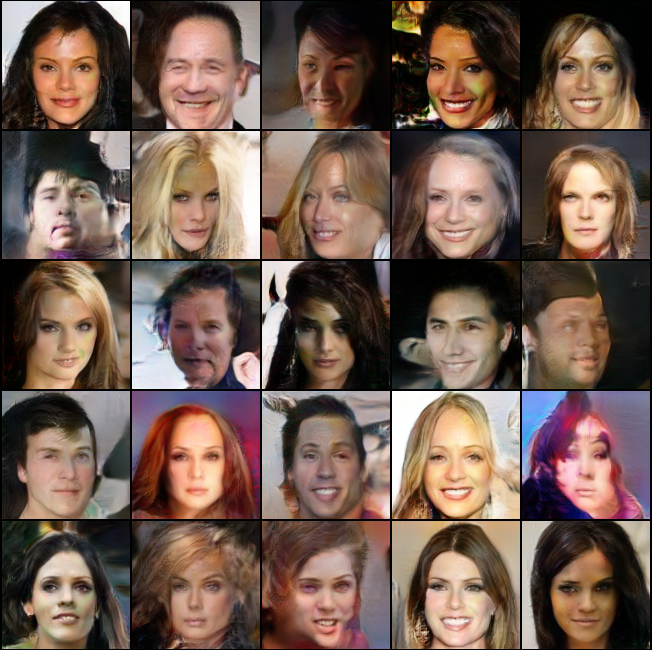}
  \caption{Randomly generated images on CelebA-128.}
  \label{fig:appceleba}
\end{figure}
In the main paper, we demonstrate experiments on $32\times32$ and $64\times64$ images, both are small scale, because of the weak expressiveness of convolutional layers based structure. To improve the expressiveness, we borrow the Resnet-based structure of SNGAN\cite{Miyato2018SpectralNF} to build our marginal EBM, inference model and IGM. And then, we use proposed compact BiDVL to train models on CIFAR-10 and CelebA-128. 

In this part, CIFAR-10 dataset consisting of $32\times32$ scale images, is to compare the generative performance of the Resnet-based model with the simple-structured model in the main paper. While the CelebA-128 dataset constructed by resizing images in CelebA to $128\times128$, is for evaluating the adaptability to higher-dimensional data. They go through the same pre-processing as in main experiments. 

We borrow the structure of the discriminator and the generator of SNGAN to build the marginal EBM and the IGM for both $32\times32$ and $128\times128$ datasets. But for the inference model, which is not included in SNGAN, we cascade spectral normalized Resblocks in the same way as in SNGAN discriminator, except that, the last linear layer is replaced with two linear layer to output the mean and the log-variance, respectively. Moreover, the log-variance is followed by a Softplus function to be bound within $(-\infty,0)$, corresponding to bound the output variance within $(0,1)$, which significantly helps the training in early stage. 

For both CIFAR-10 and CelebA-128, we apply the offset compact BiDVL (\ref{eq:pllgr},\ref{eq:pulgr}) and alternatively optimize the upper-level objective with one step and the lower-level objective with one step, \ie set the number of lower-level steps $N$ to one. 

Unfortunately, we found the complex structure exacerbates another learning problems resulting in instability of the marginal EBM. Since the output of the EBM is an unbounded real value, training with the Monte Carlo estimate of the upper-level gradient (\ref{eq:pllgr}) makes the energy easily to reduces to $-\infty$. It may because adding a constant to the whole energy landscape will not change the probability distribution:
\begin{equation*}
    p(v) = \frac{\exp(-\mathcal{E}(v))}{\int \exp(-\mathcal{E}(v)) \mathrm{d}v} = \frac{\exp(-\mathcal{E}(v) + c)}{\int \exp(-\mathcal{E}(v) + c) \mathrm{d}v},
\end{equation*}
however, largely influences the training stability. To handle the numerical problem efficiently, we turn to adopt a restricted version of \cref{eq:pllgr}:
\begin{equation}
    \begin{gathered}
        \nabla_{\psi^{\prime}} \mathcal{\widetilde{J}}_{\rm UL}(\psi^{\prime}) = \mathbb{E}_{q(v)}[\nabla_{\psi^{\prime}}\mathrm{ReLU}(1+\mathcal{E}_{\psi^{\prime}}(v))] 
        \\ +\ \mathbb{E}_{p_{\omega_2}(v,h)}[\nabla_{\psi^{\prime}}\mathrm{ReLU}(1 - \mathcal{E}_{\psi^{\prime}}(v))],
    \end{gathered}
    \label{eq:hinge}
\end{equation}
which can prevent EBM from outputting numerical unstable value. Furthermore, \cref{eq:hinge} may be regarded as a strong constraint on $\nabla_{\psi^{\prime}}\mathcal{E}_{\psi^{\prime}}(v)$ as discussed in \cref{sec:diver}, since it clips the objective directly. 

The Resnet-based model gets 16.37 FID on CIFAR-10 and demonstrates better performance than the best model shown in \cref{tab:imge1}. The generated images are presented in \cref{fig:appcifar}. For CelebA-128, we experimentally find that models simply cascading convolutional layers fails to generate meaningful images, but the Resnet-based model can generate good images as demonstrated in \cref{fig:appceleba}. Some reconstruction images are shown in \cref{fig:appcelebarecon}.

\end{document}